%% file: main.tex
\documentclass[a4paper, 10pt, conference]{ieeeconf}      % Use this line for a4 paper

\IEEEoverridecommandlockouts                              % This command is only needed if 
\usepackage{amssymb}
\usepackage{amsmath}
\usepackage[font=footnotesize,skip=4pt]{caption}
\usepackage{subcaption}
\usepackage{graphicx}      % include this line if your document contains figures
\usepackage{svg}
\usepackage{adjustbox}
\graphicspath{{./graphics/}}
\usepackage{siunitx}
\usepackage{bm}
\usepackage{algorithm}% http://ctan.org/pkg/algorithms
\usepackage{algpseudocode}% http://ctan.org/pkg/algorithmicx
\usepackage{hyperref}
\usepackage{listings}
\usepackage{tikz}
\usepackage{pgfplots}
\usepackage{xcolor}
\usepackage{ACIN_colors}
\usepackage{ACIN_macros}
\usepackage{booktabs}
\usepackage{cleveref}
\usepackage{lipsum}

\usepackage{import}  
\usepackage{svg}
\usetikzlibrary{positioning}

\pgfplotsset{compat=1.18} 

\providecommand{\kuka}{\textsc{KUKA} LBR iiwa R820\xspace}

\let\labelindent\relax
\usepackage{enumitem}
\setlist{nosep,leftmargin=*}

\newcommand{\expvideo}{\url{https://www.acin.tuwien.ac.at/42d7/}}

\title{\LARGE \bf
Autonomous Block Assembly for Boom Cranes with Passive Joint Dynamics: Integrated Vision MPC Control
}

\author{Gerald Ebmer$^{1}$, Minh Nhat Vu$^{1}$, Tobias Glück$^{2}$ and Wolfgang Kemmetmüller$^{1}$% <-this % stops a space
\thanks{$^{1}$Automation \& Control Institute (ACIN), TU Wien, Gusshausstrasse 27-29, 1040 Wien, Austria
        {\tt\small <ebmer,vu,kemmetmueller>@acin.tuwien.ac.at}}%
\thanks{$^{2}$Center for Vision, Automation \& Control, AIT Austrian Institute of Technology GmbH, Giefinggasse 4, 1210 Vienna, Austria
        {\tt\small tobias.glueck@ait.ac.at}}%
}

\begin{document}

\maketitle
\thispagestyle{empty}
\pagestyle{empty}

%%%%%%%%%%%%%%%%%%%%%%%%%%%%%%%%%%%%%%%%%%%%%%%%%%%%%%%%%%%%%%%%%%%%%%%%%%%%%%%%
\begin{abstract}
This paper presents an autonomous control framework for articulated boom cranes performing prefabricated block assembly in construction environments. The key challenge addressed is precise placement control under passive joint dynamics that cause pendulum-like sway, complicating the accurate positioning of building components. Our integrated approach combines real-time vision-based pose estimation of building blocks, collision-aware B-spline path planning, and nonlinear model predictive control (NMPC) to achieve autonomous pickup, placement, and obstacle-avoidance assembly operations.
The framework is validated on a laboratory-scale testbed that emulates crane kinematics and passive dynamics while enabling rapid experimentation. The collision-aware planner generates feasible B-spline references in real-time on CPU hardware with anytime performance, while the NMPC controller actively suppresses passive joint sway and tracks the planned trajectory under continuous vision feedback. Experimental results demonstrate autonomous block stacking and obstacle-avoidance assembly, with sway damping reducing settling times by more than an order of magnitude compared to uncontrolled passive dynamics, confirming the real-time feasibility of the integrated approach for construction automation.
% This paper presents a unified perception–planning–control framework for under-actuated manipulators with passive joints, motivated by articulated boom cranes in construction. 
% The framework integrates vision-based pose estimation, collision-free path generation, and nonlinear model predictive control (NMPC) into a closed loop that achieves real-time execution of assembly tasks. 
% A laboratory-scale testbed emulates crane kinematics and passive dynamics, enabling rapid experimentation while providing a path toward transfer to full-scale machinery. 
% The planner generates collision-free B-spline references real-time feasible on CPU hardware with anytime performance, while the NMPC stabilizes passive joint sway and advances along the path under online vision updates. 
% Experiments demonstrate autonomous block pickup, placement, and obstacle-avoidance assembly, confirming real-time feasibility and showing that sway damping reduces settling times by more than an order of magnitude. 
%The results highlight laboratory-scale testbeds as effective enablers for developing autonomy concepts for large-scale construction equipment.
\end{abstract}

 \input{tex/10_introduction}
 % \input{tex/60_integration}
 \input{tex/20_modeling}

 \input{tex/30_control}

 \input{tex/40_path_planning}
 \input{tex/54_pose_filtering}

 \input{tex/70_validation}

 \input{tex/90_conclusion}


%%%%%%%%%%%%%%%%%%%%%%%%%%%%%%%%%%%%%%%%%%%%%%%%%%%%%%%%%%%%%%%%%%%%%%%%%%%%%%%%

%%%%%%%%%%%%%%%%%%%%%%%%%%%%%%%%%%%%%%%%%%%%%%%%%%%%%%%%%%%%%%%%%%%%%%%%%%%%%%%%



% \section*{ACKNOWLEDGMENT}

% The preferred spelling of the word ÒacknowledgmentÓ in America is without an ÒeÓ after the ÒgÓ. Avoid the stilted expression, ÒOne of us (R. B. G.) thanks . . .Ó  Instead, try ÒR. B. G. thanksÓ. Put sponsor acknowledgments in the unnumbered footnote on the first page.

%%%%%%%%%%%%%%%%%%%%%%%%%%%%%%%%%%%%%%%%%%%%%%%%%%%%%%%%%%%%%%%%%%%%%%%%%%%%%%%%

\bibliographystyle{IEEEtran}
\bibliography{references.bib}

\end{document}

%% file: tex/10_introduction.tex
\section{Introduction}

The assembly of prefabricated building components is a critical bottleneck in modern construction, where heavy elements must be positioned with millimeter-level accuracy under time pressure. Articulated boom cranes are the dominant machinery for such tasks due to their large workspace and payload capacity. However, their under-actuated design—featuring passive joints that allow end-tool self-alignment while reducing actuator requirements—introduces pendulum-like sway dynamics that severely complicate precise placement operations \cite{KoG24, HeW19}.

This sway behavior is particularly problematic for autonomous construction scenarios, where human operators can no longer compensate through intuition and experience. Unlike structured manufacturing environments, construction sites present dynamic, cluttered conditions with irregular obstacles and continuously changing layouts \cite{XiC22a, GhC20, MeW20}. Robust autonomy, therefore, requires three capabilities in combination: (i) real-time vision feedback to estimate the pose of building components, (ii) collision-free motion planning in cluttered environments, and (iii) active sway damping under passive joint dynamics.

Existing research has advanced each aspect individually. Vision-based feedback has been applied in crane operations, for instance, using multi-camera systems for simultaneous payload damping and parameter estimation \cite{TyC21}, while forestry cranes have integrated stereo vision and grasp planning to achieve autonomous log loading with high success rates \cite{AyF24a}. Collision-aware trajectory optimization methods have been proposed for construction cranes, achieving near time-optimal motions under hydraulic constraints in simulation \cite{EcB25}. For sway suppression, both linear and nonlinear MPC schemes have demonstrated substantial reductions in oscillations \cite{KaB14, ScA23b}, and dynamic programming approaches achieved up to 90\% sway reduction in simulation \cite{JeS25}. Yet these advances remain fragmented: perception, planning, and control are typically addressed in isolation, and no integrated framework exists that unifies them in a closed loop for autonomous prefabricated assembly with articulated boom cranes.

\begin{figure}[t]
    \centering
    \includegraphics[width=80mm, height=45mm]{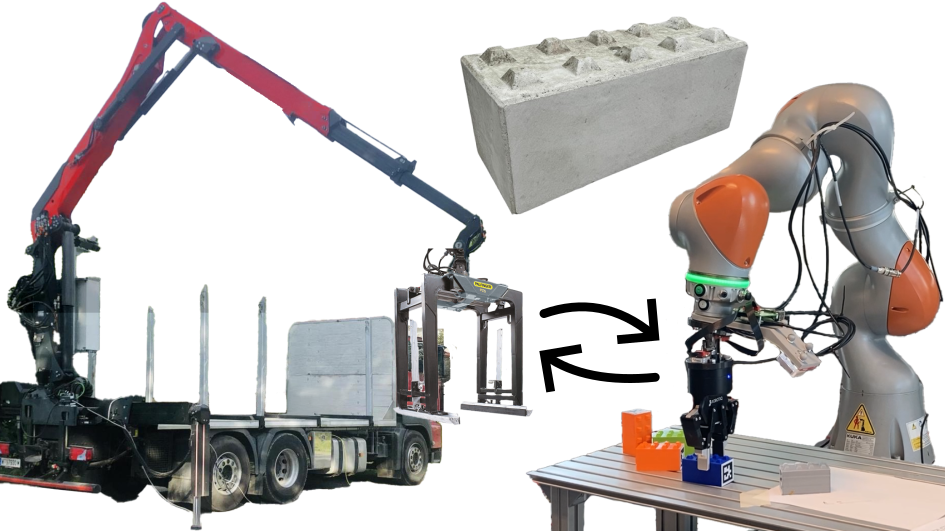}
    \caption{Motivation: articulated boom crane with concrete block (left) and developed laboratory-scale setup (right) as a testbed for autonomous assembly.}
    \label{fig:titlepage}
\end{figure}

To address this gap, we present a laboratory-scale framework that integrates real-time vision, collision-aware B-spline path generation, and nonlinear model predictive control (NMPC) into a single closed loop. The testbed emulates articulated crane kinematics and passive-joint dynamics while enabling precise performance measurement and rapid experimental cycles that would be infeasible at full scale. \Cref{fig:titlepage} illustrates the motivating crane scenario and the developed lab setup. A video of the experiments is available at \expvideo.

\subsection{Contributions}
The main contribution of this work is a unified framework and laboratory testbed for developing autonomous control of large-scale, under-actuated articulated boom cranes. Specifically:
\begin{itemize}
    \item \textbf{Integrated perception–planning–control architecture:} Real-time vision feedback, collision-free path generation, and nonlinear MPC are combined in a single closed loop that accounts for passive joint dynamics and provides active sway damping.  
    \item \textbf{Laboratory-scale validation:} A physical testbed emulating articulated crane dynamics is used to demonstrate autonomous pick-and-place and obstacle-avoidance assembly. The experiments confirm the framework’s real-time feasibility and ability to suppress passive-joint sway by more than an order of magnitude, establishing a foundation for transfer to full-scale machinery.  
\end{itemize}

\subsection{Paper's Structure}
\Cref{sec:modelling} presents the dynamic model of the lab setup and \Cref{sec:control} discusses the nonlinear MPC formulation.  
\Cref{sec:path_planning} presents the path planner, followed in \Cref{sec:pose_estimation} by the vision-based pose estimation.
\Cref{sec:laboratory_setup} details the lab setup and framework integration, followed by the results in \Cref{sec:results} and the conclusion in \Cref{sec:conclusion}.

% \Cref{sec:modelling} presents the dynamic model of the lab setup emulating articulated boom cranes.
% \Cref{sec:control} formulates the nonlinear MPC for path following and sway damping.  
% \Cref{sec:path_planning} presents the stochastic sampling-based planner for collision-free B-spline paths, followed in \Cref{sec:pose_estimation} by the RGB/Aruco-based vision pipeline with filtering for online pose updates.  
% \Cref{sec:laboratory_setup} details the lab-scale setup, including low-level robot control and integration with the MPC.  
% \Cref{sec:results} reports experimental block assembly and quantifies the sway damping achieved by the controller.  
% Finally, \Cref{sec:conclusion} summarizes the contributions and discusses the scaling to full-size machinery.

%% file: tex/20_modeling.tex
\section{Modeling}
\label{sec:modelling}
\begin{figure*}[t]
    \centering
    \def\svgwidth{\textwidth} % relative width
    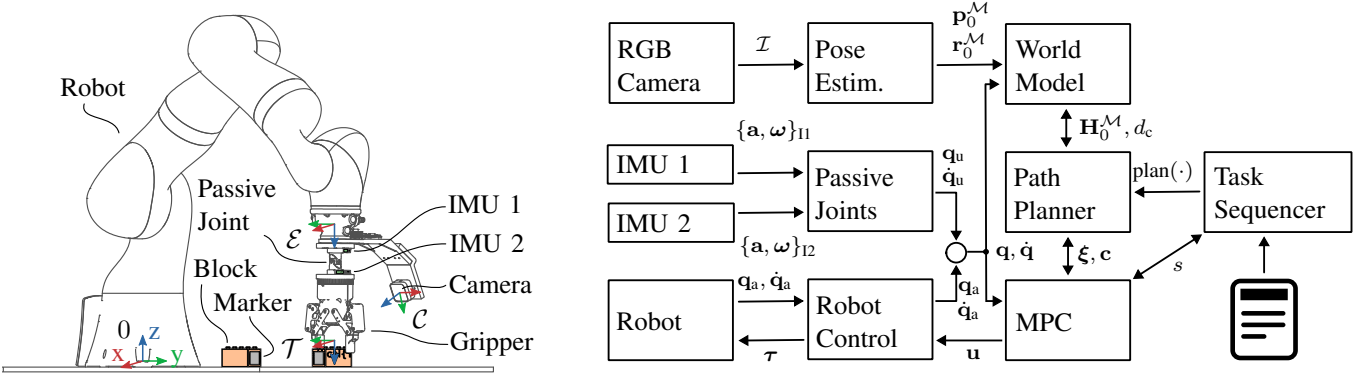
    \caption{System overview. Left: schematic of the laboratory setup with robot, passive joint, gripper, blocks, and reference frames ${0}$ (base), $\mathcal{E}$ (end-effector), $\mathcal{T}$ (tool center), and $\mathcal{C}$ (camera). Right: block diagram of the integrated framework, illustrating the interconnection of perception, path planning, MPC control, and the environment model. The task sequencer executes pickup and placement routines based on a configuration file specifying the schedule.}
    \label{fig:system_overview}
\end{figure*}

The considered 9-DoF system is depicted on the left side of \Cref{fig:system_overview}. 
It consists of the lightweight industrial robot \kuka with seven actuated DoFs, a cardan joint (passive joint) comprised of two non-actuated DoFs, and the Robotiq 2F-85 gripper. 
The design of the non-actuated joints resembles a forestry crane \cite{EcB25} equipped with a passive joint.

%, which is commonly used to reduce stress on the mechanical construction induced by the load. 

The equations of motion of the 9-DoF system with the generalized coordinates $\transpose{\vec{q}} = {[q_1, \dots, q_9]}$ can be written as
\begin{equation}
    \vec{M}(\vec{q}) \Ddot{\vec{q}} + \vec{C}(\vec{q},\dot{\vec{q}}) \dot{\vec{q}} + \vec{g}(\vec{q}) + \vec{D} \dot{\vec{q}} = \vec{\tau}
    \label{eq:eq_of_motion}
\end{equation}
with the mass matrix $\vec{M}(\vec{q})$, the Coriolis matrix $\vec{C}(\vec{q},\dot{\vec{q}})$, the gravitational forces $\vec{g}(\vec{q})$, the viscous friction matrix  $\vec{D}$, and the generalized forces $\vec{\tau}$ \cite{SiS09, LyP17}.
For further consideration, the nine DoFs are split up into actuated $\transpose{\vec{q}_\text{a}} = {[q_1, \dots, q_7]}$ and non-actuated $\transpose{\vec{q}_\text{u}} = {[q_8, q_9]}$ DoFs. 
Based on this separation, \eqref{eq:eq_of_motion} can be rearranged as
\begin{align}
    &
    \begin{bmatrix}
        \vec{M}_\text{a} & \vec{M}_\text{au} \\
        \vec{M}_\text{ua}& \vec{M}_\text{u} 
    \end{bmatrix}
    \begin{bmatrix}
        \Ddot{\vec{q}}_\text{a} \\
        \Ddot{\vec{q}}_\text{u}
    \end{bmatrix}
    +
    \begin{bmatrix}
        \vec{C}_\text{a} & \vec{C}_\text{au} \\
        \vec{C}_\text{ua} & \vec{C}_\text{u}
    \end{bmatrix}
    \begin{bmatrix}
        \dot{\vec{q}}_\text{a} \\
        \dot{\vec{q}}_\text{u}
    \end{bmatrix}
    +
    \begin{bmatrix}
        {\vec{g}}_\text{a} \\
        {\vec{g}}_\text{u}
    \end{bmatrix} 
    \nonumber \\
    &
    +
    \begin{bmatrix}
        \vec{D}_\text{a} & \vec{0} \\
        \vec{0} & \vec{D}_\text{u}
    \end{bmatrix}
    \begin{bmatrix}
        \dot{\vec{q}}_\text{a} \\
        \dot{\vec{q}}_\text{u}
    \end{bmatrix}
    =
    \begin{bmatrix}
        {\vec{\tau}}_\text{a} \\
        {\vec{0}}
    \end{bmatrix}
    \FullStop
    \label{eq:split_system}
\end{align}
Considering the bottom line of \eqref{eq:split_system}, the dynamics of the non-actuated system can be represented as
\begin{equation}
    \Ddot{\vec{q}}_\text{u} = {\vec{M}}_\text{u}^{-1} \left( -{\vec{M}}_\text{ua} \Ddot{\vec{q}}_\text{a} - {\vec{C}}_\text{ua} \dot{\vec{q}}_\text{a} - {\vec{C}}_\text{u} \dot{\vec{q}}_\text{u} - {\vec{g}}_\text{u} - {\vec{D}}_\text{u} \dot{\vec{q}}_\text{u} \right) \FullStop
    \label{eq:full_ua_dyn}
\end{equation}
To reduce the complexity of the model \eqref{eq:full_ua_dyn}, the Coriolis terms $\vec{C}_\text{ua} \dot{\vec{q}}_\text{a}$ and $\vec{C}_\text{u} \dot{\vec{q}}_\text{u}$ are neglected as their contribution to the system dynamics is minor; see \eg \cite{VuH21b}. This yields the simplified dynamics of the non-actuated system
\begin{equation}
    \Ddot{\vec{q}}_\text{u} = {\vec{M}}_\text{u}^{-1} \left( -{\vec{M}}_\text{ua} \Ddot{\vec{q}}_\text{a} - {\vec{g}}_\text{u} - {\vec{D}}_\text{u} \dot{\vec{q}}_\text{u} \right) \FullStop
    \label{eq:ua_dyn}
\end{equation}

It is further assumed that the actuated joints are controlled by a sub-ordinate controller such that the system input $\vec{u}$ for the nonlinear MPC can be chosen as $\vec{u} = \ddot{\vec{q}}_\text{a}$. The resulting system dynamics are
\begin{equation}
\dot{\vec{x}} = \vec{f}(\vec{x},\vec{u}) = 
    \begin{bmatrix}
        {\vec{\omega}}_\text{a} \\
        {\vec{\omega}}_\text{u} \\
        \vec{u} \\
        {\vec{M}}_\text{u}^{-1} \left( -{\vec{M}}_\text{ua} {\vec{u}} - {\vec{g}}_\text{u} - {\vec{D}}_\text{u} {\vec{\omega}}_\text{u} \right)
    \end{bmatrix} \Comma
\label{eq:sysdyn}
\end{equation}
with the system state $\transpose{\vec{x}} = [\transpose{\vec{q}}, \transpose{{\vec{\omega}}}]$ and $\vec{\omega}=\dot{\vec{q}}$.

%% file: graphics/overview/overview_tex.pdf_tex
%% Creator: Inkscape 1.4.2 (f4327f4, 2025-05-13), www.inkscape.org
%% Accompanies image file 'overview_tex.pdf' (pdf, eps, ps)
\begingroup%
  \makeatletter%
  \providecommand\color[2][]{%
    \errmessage{(Inkscape) Color is used for the text in Inkscape, but the package 'color.sty' is not loaded}%
    \renewcommand\color[2][]{}%
  }%
  \providecommand\transparent[1]{%
    \errmessage{(Inkscape) Transparency is used (non-zero) for the text in Inkscape, but the package 'transparent.sty' is not loaded}%
    \renewcommand\transparent[1]{}%
  }%
  \providecommand\rotatebox[2]{#2}%
  \newcommand*\fsize{\dimexpr\f@size pt\relax}%
  \newcommand*\lineheight[1]{\fontsize{\fsize}{#1\fsize}\selectfont}%
  \ifx\svgwidth\undefined%
    \setlength{\unitlength}{510.23622047bp}%
    \ifx\svgscale\undefined%
      \relax%
    \else%
      \setlength{\unitlength}{\unitlength * \real{\svgscale}}%
    \fi%
  \else%
    \setlength{\unitlength}{\svgwidth}%
  \fi%
  \global\let\svgwidth\undefined%
  \global\let\svgscale\undefined%
  \makeatother%
  \begin{picture}(1,0.27777778)%
    \lineheight{1}%
    \setlength\tabcolsep{0pt}%
    \put(0,0){\includegraphics[width=\unitlength,page=1]{overview_tex.pdf}}%
    \put(0.10792149,0.02867132){\color[rgb]{0.17254902,0.40392157,0.64313725}\makebox(0,0)[lt]{\lineheight{1.25}\smash{\begin{tabular}[t]{l}z\end{tabular}}}}%
    \put(0.12423961,0.0116779){\color[rgb]{0.04705882,0.67843137,0.25882353}\makebox(0,0)[lt]{\lineheight{1.25}\smash{\begin{tabular}[t]{l}y\end{tabular}}}}%
    \put(0.08057141,0.0126094){\color[rgb]{0.77254902,0.2,0.21176471}\makebox(0,0)[lt]{\lineheight{1.25}\smash{\begin{tabular}[t]{l}x\end{tabular}}}}%
    \put(0,0){\includegraphics[width=\unitlength,page=2]{overview_tex.pdf}}%
    \put(0.20608705,0.0158151){\color[rgb]{0,0,0}\makebox(0,0)[lt]{\lineheight{1.25}\smash{\begin{tabular}[t]{l}$\mathcal{T}$\end{tabular}}}}%
    \put(0,0){\includegraphics[width=\unitlength,page=3]{overview_tex.pdf}}%
    \put(0.30339726,0.03751628){\color[rgb]{0,0,0}\makebox(0,0)[lt]{\lineheight{1.25}\smash{\begin{tabular}[t]{l}$\mathcal{C}$\end{tabular}}}}%
    \put(0,0){\includegraphics[width=\unitlength,page=4]{overview_tex.pdf}}%
    \put(0.20966933,0.0966593){\color[rgb]{0,0,0}\makebox(0,0)[lt]{\lineheight{1.25}\smash{\begin{tabular}[t]{l}$\mathcal{E}$\end{tabular}}}}%
    \put(0,0){\includegraphics[width=\unitlength,page=5]{overview_tex.pdf}}%
    \put(0.08494306,0.02975472){\color[rgb]{0,0,0}\makebox(0,0)[lt]{\lineheight{1.25}\smash{\begin{tabular}[t]{l}$0$\end{tabular}}}}%
    \put(0.04267585,0.21040911){\makebox(0,0)[lt]{\lineheight{1.25}\smash{\begin{tabular}[t]{l}Robot\end{tabular}}}}%
    \put(0.15588462,0.0501773){\makebox(0,0)[lt]{\lineheight{1.25}\smash{\begin{tabular}[t]{l}Marker\end{tabular}}}}%
    \put(0.33023237,0.06363871){\makebox(0,0)[lt]{\lineheight{1.25}\smash{\begin{tabular}[t]{l}Camera\end{tabular}}}}%
    \put(0.14118554,0.07516573){\makebox(0,0)[lt]{\lineheight{1.25}\smash{\begin{tabular}[t]{l}Block\end{tabular}}}}%
    \put(0.33116171,0.02097598){\makebox(0,0)[lt]{\lineheight{1.25}\smash{\begin{tabular}[t]{l}Gripper\end{tabular}}}}%
    \put(0,0){\includegraphics[width=\unitlength,page=6]{overview_tex.pdf}}%
    \put(0.32949545,0.12309588){\makebox(0,0)[lt]{\lineheight{1.25}\smash{\begin{tabular}[t]{l}IMU 1\end{tabular}}}}%
    \put(0.3308036,0.09411774){\makebox(0,0)[lt]{\lineheight{1.25}\smash{\begin{tabular}[t]{l}IMU 2\end{tabular}}}}%
    \put(0.14423648,0.13482155){\makebox(0,0)[lt]{\lineheight{1.25}\smash{\begin{tabular}[t]{l}Passive\\Joint\end{tabular}}}}%
    \put(0,0){\includegraphics[width=\unitlength,page=7]{overview_tex.pdf}}%
    \put(0.45501275,0.23873709){\makebox(0,0)[lt]{\lineheight{1.25}\smash{\begin{tabular}[t]{l}RGB\\Camera\end{tabular}}}}%
    \put(0,0){\includegraphics[width=\unitlength,page=8]{overview_tex.pdf}}%
    \put(0.60126656,0.23873709){\makebox(0,0)[lt]{\lineheight{1.25}\smash{\begin{tabular}[t]{l}Pose\\Estim.\end{tabular}}}}%
    \put(0.74946459,0.23873709){\makebox(0,0)[lt]{\lineheight{1.25}\smash{\begin{tabular}[t]{l}World\\Model\end{tabular}}}}%
    \put(0.45427581,0.15291404){\makebox(0,0)[lt]{\lineheight{1.25}\smash{\begin{tabular}[t]{l}IMU 1\end{tabular}}}}%
    \put(0.45427581,0.10881681){\makebox(0,0)[lt]{\lineheight{1.25}\smash{\begin{tabular}[t]{l}IMU 2\end{tabular}}}}%
    \put(0.60200348,0.04764911){\makebox(0,0)[lt]{\lineheight{1.25}\smash{\begin{tabular}[t]{l}Robot\\Control\end{tabular}}}}%
    \put(0.45427581,0.03539984){\makebox(0,0)[lt]{\lineheight{1.25}\smash{\begin{tabular}[t]{l}Robot\end{tabular}}}}%
    \put(0.74965269,0.03524303){\makebox(0,0)[lt]{\lineheight{1.25}\smash{\begin{tabular}[t]{l}MPC\end{tabular}}}}%
    \put(0.89637689,0.14388294){\makebox(0,0)[lt]{\lineheight{1.25}\smash{\begin{tabular}[t]{l}Task\\Sequencer\end{tabular}}}}%
    \put(0.74825728,0.1431931){\makebox(0,0)[lt]{\lineheight{1.25}\smash{\begin{tabular}[t]{l}Path\\Planner\end{tabular}}}}%
    \put(0.60126656,0.1431931){\makebox(0,0)[lt]{\lineheight{1.25}\smash{\begin{tabular}[t]{l}Passive\\Joints\end{tabular}}}}%
    \put(0,0){\includegraphics[width=\unitlength,page=9]{overview_tex.pdf}}%
    \put(0.54367972,0.06825456){\makebox(0,0)[lt]{\lineheight{1.25}\smash{\begin{tabular}[t]{l}\footnotesize $\vec{q}_\text{a},\dot{\vec{q}}_\text{a}$\end{tabular}}}}%
    \put(0.73493076,0.09069922){\makebox(0,0)[lt]{\lineheight{1.25}\smash{\begin{tabular}[t]{l}\footnotesize $\vec{q}, \dot{\vec{q}}$\end{tabular}}}}%
    \put(0.56286473,0.01293962){\makebox(0,0)[lt]{\lineheight{1.25}\smash{\begin{tabular}[t]{l}\footnotesize $\vec{\tau}$\end{tabular}}}}%
    \put(0.70250596,0.26630858){\makebox(0,0)[lt]{\lineheight{1.25}\smash{\begin{tabular}[t]{l}\footnotesize $\vec{p}_{0}^{\mathcal{M}}$\end{tabular}}}}%
    \put(0.70250596,0.24371512){\makebox(0,0)[lt]{\lineheight{1.25}\smash{\begin{tabular}[t]{l}\footnotesize $\vec{r}_{0}^{\mathcal{M}}$\end{tabular}}}}%
    \put(0.86584396,0.079744){\makebox(0,0)[lt]{\lineheight{1.25}\smash{\begin{tabular}[t]{l}\footnotesize $s$\end{tabular}}}}%
    \put(0.71268768,0.01408028){\makebox(0,0)[lt]{\lineheight{1.25}\smash{\begin{tabular}[t]{l}\footnotesize $\vec{u}$\end{tabular}}}}%
    \put(0,0){\includegraphics[width=\unitlength,page=10]{overview_tex.pdf}}%
    \put(0.69515639,0.16341506){\makebox(0,0)[lt]{\lineheight{1.25}\smash{\begin{tabular}[t]{l}\footnotesize $\vec{q}_\text{u}$\end{tabular}}}}%
    \put(0.69515639,0.14577617){\makebox(0,0)[lt]{\lineheight{1.25}\smash{\begin{tabular}[t]{l}\footnotesize $\dot{\vec{q}}_\text{u}$\end{tabular}}}}%
    \put(0.70710322,0.06450422){\makebox(0,0)[lt]{\lineheight{1.25}\smash{\begin{tabular}[t]{l}\footnotesize $\vec{q}_\text{a}$\end{tabular}}}}%
    \put(0.70644781,0.04582246){\makebox(0,0)[lt]{\lineheight{1.25}\smash{\begin{tabular}[t]{l}\footnotesize $\dot{\vec{q}}_\text{a}$\end{tabular}}}}%
    \put(0.79657974,0.18197901){\makebox(0,0)[lt]{\lineheight{1.25}\smash{\begin{tabular}[t]{l}\footnotesize $\vec{H}_{0}^\mathcal{M}, d_\text{c}$\end{tabular}}}}%
    \put(0.83626709,0.14964105){\makebox(0,0)[lt]{\lineheight{1.25}\smash{\begin{tabular}[t]{l}\footnotesize $\text{plan}(\cdot)$\end{tabular}}}}%
    \put(0,0){\includegraphics[width=\unitlength,page=11]{overview_tex.pdf}}%
    \put(0.55774918,0.24371512){\makebox(0,0)[lt]{\lineheight{1.25}\smash{\begin{tabular}[t]{l}\footnotesize $\mathcal{I}$\end{tabular}}}}%
    \put(0.54228609,0.18050912){\makebox(0,0)[lt]{\lineheight{1.25}\smash{\begin{tabular}[t]{l}\footnotesize $\{ \vec{a}, \vec{\omega}\}_\text{I1}$\end{tabular}}}}%
    \put(0.54522591,0.09378457){\makebox(0,0)[lt]{\lineheight{1.25}\smash{\begin{tabular}[t]{l}\footnotesize $\{ \vec{a}, \vec{\omega}\}_\text{I2}$\end{tabular}}}}%
    \put(0.79546678,0.0877594){\makebox(0,0)[lt]{\lineheight{1.25}\smash{\begin{tabular}[t]{l}\footnotesize $\vec{\xi}, \vec{c}$\end{tabular}}}}%
  \end{picture}%
\endgroup%

%% file: tex/30_control.tex
\section{Path-Following MPC Incorporating Passive Joint Dynamics}
\label{sec:control}

The path-following MPC approach is formulated as a finite-horizon optimal control problem that optimizes the path parameter progression and the system's motion along a predefined spline-based path. Due to the under-actuated nature of the system, only the gripper's position and yaw angle can be directly controlled, necessitating a 4D task space representation $\vec{\xi} = \transvec{x, y, z, \psi}$ where the passive joint dynamics determine roll and pitch orientations. 
The MPC formulation incorporates the passive joint dynamics while advancing the path to enable precise assembly operations.

\subsection{Spline-Based Path Parameterization}
Task space path parameterization is preferred for assembly operations as it provides smoother end-effector trajectories than joint space paths, which introduce irregular Cartesian motions due to kinematic nonlinearities. The path parameter $s \in [0,1]$ represents progression along the 4D path defined by a B-spline $\vec{\xi}_{\text{d}}(s) = \sum_{i=0}^{n_\text{cp}-1} \vec{c}_{i} B_{i,k}(s)$ from a start pose $\vec{\xi}_\text{s} = \vec{\xi}_\text{d}(0)$ to an end pose $\vec{\xi}_\text{e} = \vec{\xi}_\text{d}(1)$, where $B_{i,k}(s)$ are B-spline basis functions of degree $k$, and each control point $\vec{c}_{i} = \transvec{x_i, y_i, z_i, \psi_i}$, $i=0, \dots, n_\text{cp}-1$ contains position and yaw information.

\subsection{Inverse Kinematics and Steady State of the Under-actuated System}
Due to the under-actuated dynamics, inverse kinematics solutions for a stationary end pose $\vec{\xi}_\text{e} = \transvec{\transpose{\vec{p}}_\text{e}, \psi_\text{e}}$ must correspond to a steady state of the non-actuated subsystem to avoid unwanted sway motions. At steady state, both accelerations $\ddot{\vec{q}}_\text{u}$ and velocities $\dot{\vec{q}}_\text{u}$ of the under-actuated joints are zero.
From \eqref{eq:ua_dyn}, the steady-state condition requires $\vec{g}_\text{u}(\vec{q}) = \vec{0}$, where $\vec{g}_\text{u}(\vec{q})$ represents the gravitational forces acting on the under-actuated joints.
Using the error formulation similar to \cite{Hu09}, the steady-state configuration $\vec{q}_\text{e}$ is obtained by solving
\begin{align}
\label{eq:ik}
    \vec{q}_\text{e} = \arg \min_{\vec{q}} \; & \Big[ \norm{ \vec{p}(\vec{q}) - \vec{p}_\text{e} }_2^2 
    + %w_{\text{rot}} 
    \norm{ \transpose{\vec{R}(\psi_\text{e})} \vec{R}(\vec{q}) - \mathbf{I} }_\text{F}^2 \Big] \\
    & \text{subject to} \quad \vec{g}_\text{u}(\vec{q}) = \vec{0} \nonumber
\end{align}
where $\norm{\cdot}_\text{F}$ denotes the Frobenius norm.
%, $\mathbf{I}$ is the identity matrix, and $w_{\text{rot}} = 0.1$ is chosen empirically. The orientation error formulation follows 
The optimization problem \eqref{eq:ik} is solved with Ipopt \cite{WaB06a}.

\subsection{Optimal Control Problem Formulation}
Our control strategy aims are twofold: Firstly, it should incorporate the under-actuated system dynamics while stabilizing the gripper, and secondly, it should progress the path. 
For this we augment the original system state $\vec{x}$ with the path parameter $s$ and its velocity $\dot{s}$ resulting the augmented system state
$ \bar{\vec{x}} = \transvec{ \transpose{\vec{x}}, s, \dot{s}} \in \mathbb{R}^{2n_q + 2}$, 
as well as the augmented system input comprising the acceleration of the seven actuated joints and the path parameter's second time derivative as input
$ \bar{\vec{u}} = \transvec{ \transpose{\vec{u}}, v} = \transvec{ \transpose{\ddot{\Vec{q}}_{\text{a}}}, \ddot{s}} \in \mathbb{R}^{n_{\text{a}} + 1}$

Based on \eqref{eq:sysdyn}, the augmented system dynamics result in
\begin{equation}
    \dot{\bar{\vec{x}}} = \bar{\vec{f}}(\bar{\vec{x}}, \bar{\vec{u}}) = \begin{bmatrix}
        \vec{f}(\vec{x},\vec{u}) \\
        \dot{s} \\
        v
    \end{bmatrix} \FullStop
    \label{eq:aug_dynamics}
\end{equation}

The optimal control problem for the path-following MPC is defined as
\begin{subequations}
    \begin{alignat}{2}
        &\min_{\substack{\bar{\vec{x}}_{1|n}, \dots, \bar{\vec{x}}_{N|n} \\ 
            \bar{\vec{u}}_{1|n}, \dots, \bar{\vec{u}}_{N-1|n}}} 
            &&\sum_{k=0}^{N-1} l \left(\bar{\vec{x}}_{k|n}, \bar{\vec{u}}_{k|n} \right) + m \left(\bar{\vec{x}}_{N|n} \right) \label{eq:ocp} \\
        & \quad \text{subject to} \ &&\bar{\vec{x}}_{k+1|n} = \vec{F}(\bar{\vec{x}}_{k|n}, \bar{\vec{u}}_{k|n})  \label{eq:sys_constr}\\
        & \quad && \bar{\vec{x}}_{0|n} = \bar{\vec{x}}(t_n) \label{eq:ocp_init} \\
        & \quad && \bar{\vec{x}}^\text{-} \leq \bar{\vec{x}}_{k|n} \leq \bar{\vec{x}}^\text{+} ,\quad \bar{\vec{u}}^\text{-} \leq \bar{\vec{u}}_{k|n} \leq \bar{\vec{u}}^\text{+} 
        \label{eq:ocp_limit}
    \end{alignat}
    \label{eq:ocp_formulation}
\end{subequations}
where $n$ indicates the iteration of the OCP and $k=0, \ldots, N$ refers to the collocation points over the prediction horizon with length $N$. The discrete-time system dynamics $\bar{\vec{x}}_{k+1|n} = \Vec{F}(\cdot)$ are obtained through implicit Runge-Kutta integration of \eqref{eq:aug_dynamics} with the discretization timestep $T_\text{h}$.
The state limits $\bar{\vec{x}}^\pm = \transvec{{\vec{q}}^{\pm}_\text{a}, {\vec{q}}^{\pm}_\text{u}, {\dot{\vec{q}}}^{\pm}_\text{a}, {\dot{\vec{q}}}^{\pm}_\text{u}, s^{\pm}, \dot{s}^{\pm}}$ and input limits $\bar{\vec{u}}^\pm = \transvec{\vec{u}^\pm, v^\pm} $ in \eqref{eq:ocp_limit} ensure the physical feasibility of the motion by including the joint limits and velocity limits for the robot
${\vec{q}}^\pm_\text{a} = \pm \transvec{162, 114, 162, 114, 162, 114, 166}$ in \si{\degree}, 
${\vec{q}}^\pm_\text{u} = \pm \transvec{43, 43}$ in \si{\degree} and
${\dot{\vec{q}}}^\pm_\text{a} = \pm \transvec{85, 85, 100, 75, 130, 135, 135}$ in \si{\degree\per\second}.
The remaining parameters were chosen empirically to achieve good performance
in the experiments reported in \Cref{sec:results} 
${\dot{\vec{q}}}^\pm_\text{u} = \pm \transvec{50, 50}$ in \si{\degree\per\second},
$s^\text{+} = 1$, $s^\text{-} = 0$, 
$\dot{s}^\pm = \pm 1$ in \si{\per\second} 
and $\vec{u}^\pm = \pm \vec{1}$ in \si{\radian\per\second\squared}, 
$v^\pm = \pm 1$ in \si{\per\second\squared}, where \(\vec 1\) denotes an all-ones vector of appropriate dimension.

% The cost function in \eqref{eq:ocp} is divided into the Lagrange cost $l \left(\bar{\vec{x}}_{k|n}, \bar{\vec{u}}_{k|n} \right)$ and the Mayer cost term $ m \left(\bar{\vec{x}}_{N|n} \right)$. 
The Lagrange cost $l \left(\bar{\vec{x}}_{k|n}, \bar{\vec{u}}_{k|n} \right)$ of the cost function \eqref{eq:ocp} consists of task space tracking, joint space regularization, and control effort minimization part in the form of
% \begin{align}
%     l(\bar{\vec{x}}_{k|n}, \bar{\vec{u}}_{k|n}) &= \|\tilde{\vec{p}}(\vec{q}_{k|n})\|^2_{\vec{Q}_{\text{pos}}} + \|\tilde{\psi}(\vec{q}_{k|n})\|^2_{Q_{\text{rot}}} \nonumber \\
%     &\quad + \|\tilde{\vec{q}}_{k|n}\|^2_{\vec{Q}_{\vec{q}}} + \|\tilde{\dot{\vec{q}}}_{k|n}\|^2_{\vec{Q}_{\dot{\vec{q}}}} \nonumber \\
%     &\quad + \|\tilde{s}_{k|n}\|^2_{Q_s} + \|\tilde{\dot{s}}_{k|n}\|^2_{Q_{\dot{s}}} \nonumber \\
%     &\quad + \|\vec{u}_{k|n}\|^2_{\vec{R}_{\vec{u}}} + \|v_{k|n}\|^2_{R_v}
%     \label{eq:lagrange_cost}
% \end{align}
\begin{align}
    l(\bar{\vec{x}}_{k|n}, \bar{\vec{u}}_{k|n}) &= \|\tilde{\vec{p}}(\vec{q}_{k|n})\|^2_{\vec{Q}_{\text{pos}}} + \|\tilde{\psi}(\vec{q}_{k|n})\|^2_{Q_{\text{rot}}} \nonumber \\
    &\quad + \|\tilde{\vec{q}}_{k|n}\|^2_{\vec{Q}_{q}} + \|\tilde{\dot{\vec{q}}}_{k|n}\|^2_{\vec{Q}_{\dot{q}}} \nonumber \\
    &\quad + \|\tilde{s}_{k|n}\|^2_{\eta Q_s} + \|\tilde{\dot{s}}_{k|n}\|^2_{Q_{\dot{s}}} \nonumber \\
    &\quad + \|\vec{u}_{k|n}\|^2_{\vec{R}_{u}} + \|v_{k|n}\|^2_{R_v}
    \label{eq:lagrange_cost}
\end{align}
with weighted norms $\norm{\cdot}_Q$. 
The task space errors $ \tilde{\vec{p}}(\vec{q}_{k|n}) = \vec{p}(\vec{q}_{k|n}) - \vec{p}_{\text{d}}(s_{k|n})$ and $\tilde{\psi}(\vec{q}_{k|n}) = \psi(\vec{q}_{k|n}) - \psi_{\text{d}}(s_{k|n})$ with the Cartesian position $\vec{p}(\vec{q}_{k|n})$ and the gripper's yaw angle $\psi(\vec{q}_{k|n})$ depend on the joint configuration through forward kinematics, while $\vec{p}_{\text{d}}(s_{k|n})$, and $\psi_{\text{d}}(s_{k|n})$ are obtained from the evaluation of the reference path parameterized as B-spline.
% The joint space and path parameter errors are defined similarily as $\tilde{\vec{q}}_{k|n} = \vec{q}_{k|n} - \vec{q}_{\text{d}}(s_{k|n})$, $\tilde{\dot{\vec{q}}}_{k|n} = \dot{\vec{q}}_{k|n} - \dot{\vec{q}}_{\text{d}}(s_{k|n})$
% and $\tilde{s}_{k|n} = s_{k|n} - s_{\text{d}}$, $\tilde{\dot{s}}_{k|n} = \dot{s}_{k|n} - \dot{s}_{\text{d}}$,
The joint-space regularization terms are defined relative to a nominal posture $\vec{q}_\text{nom}$ (typically $\vec{q}_\text{nom} = \vec{q}_\text{e}$) and zero velocity as 
$\tilde{\vec{q}}_{k|n} = \vec{q}_{k|n} - \vec{q}_\text{nom}$, 
$\tilde{\dot{\vec{q}}}_{k|n} = \dot{\vec{q}}_{k|n}$. 
The path-parameter tracking error is 
$\tilde{s}_{k|n} = s_{k|n} - s_\text{d}$ with regularization 
$\tilde{\dot{s}}_{k|n} = \dot{s}_{k|n}$.
The path parameter weight $Q_s$ is scaled by $\eta = 1/{\max(\|\vec{p}_\text{e} - \vec{p}_\text{s}\|_2, 0.1)}$ where $\vec{p}_\text{s}$ and $\vec{p}_\text{e}$ denote the path's start and end position, respectively. 

The Mayer cost term at the final stage $m \left(\bar{\vec{x}}_{N|n} \right)$ is similar to the Lagrange cost \eqref{eq:lagrange_cost}, except that it excludes $\|\vec{u}_{k|n}\|^2_{\vec{R}_{u}}$ and $\|v_{k|n}\|^2_{R_v}$.
% without dependencies on the input $\bar{\vec{u}}$
% \begin{align}
%     m(\bar{\vec{x}}_{N|n}) &= \|\tilde{\vec{p}}(\vec{q}_{N|n})\|^2_{\vec{Q}_{e,\text{pos}}} + \|\tilde{\psi}(\vec{q}_{N|n})\|^2_{Q_{e,\text{rot}}} \nonumber \\
%     &\quad + \|\tilde{\vec{q}}_{N|n}\|^2_{\vec{Q}_{e,\vec{q}}} + \|\tilde{\dot{\vec{q}}}_{N|n}\|^2_{\vec{Q}_{e,\dot{\vec{q}}}} \nonumber \\
%     &\quad + \|\tilde{s}_{N|n}\|^2_{Q_{e,s}} + \|\tilde{\dot{s}}_{N|n}\|^2_{Q_{e,\dot{s}}} \FullStop
%     \label{eq:mayer_cost}
% \end{align}
The values of the weight matrices and MPC parameters are listed in \Cref{sec:results}.

% Kinematic redundancy is resolved by penalizing deviations of non-crane-like joints with high weights in $\vec{Q}_{\vec{q}}$.

Kinematic redundancy is resolved by regularizing joints toward a nominal posture $\vec q_\text{nom}$ with large weights in $\mathbf{Q}_q$ assigned to joints that do not exist in articulated boom cranes (\eg, the robot's axial rotations $q_2$ and $q_6$), while $\mathbf{Q}_{\dot q}$ damps joint velocities. 
%We do not use a path-dependent joint reference $\vec q_\text{d}(s)$; tracking is performed purely in task space $(x,y,z,\psi)$.

% The kinematic redundancy of the 9-DOF manipulator for the 4-DOF task requires management of nullspace motions. 
% Specific joints not present in typical articulated crane kinematics are assigned reference values $\vec{q}_{\text{d}}$ with correspondingly high weights, effectively removing redundancy by constraining these degrees of freedom. 

\subsection{Real-Time Implementation}
The computational efficiency required for real-time control is achieved through several key implementation choices. Traditional Euler-Lagrange formulations to obtain the equations of motion result in large symbolic expressions that lead to prohibitively long compile times and, most importantly, high computational cost, often producing hundreds of megabytes of generated code, particularly for computing the computationally expensive Hessian matrix.
To address these challenges, the implementation utilizes the Recursive Newton-Euler Algorithm (RNEA) \cite{Fe08} as implemented in the Pinocchio library \cite{CaS19a}. This approach, combined with CasADi's automatic differentiation capabilities \cite{AnG19}, enables efficient computation of the robot dynamics and their derivatives. The seamless integration of CasADi with the ACADOS optimal control framework \cite{VeF22b} allows for automatic generation, compilation, and execution of optimized solver binaries.
The B-spline path parameterization used in \eqref{eq:lagrange_cost} is implemented using a custom-made CasADi-compatible library, ensuring efficient evaluation of reference trajectories and their derivatives within the optimization framework. 

The ACADOS-generated solver employs the Real-Time Iteration (RTI) algorithm \cite{diehl2005real} optimized for real-time performance, using a single QP iteration per MPC cycle with HPIPM as the QP solver featuring partial condensing and a maximum of 20 internal iterations \cite{FrD20a}. The configuration includes warm starting for accelerated convergence, merit function backtracking for robust convergence, exact Hessian computation, and implicit Runge-Kutta (IRK) integration for numerical stability. The computation times of \eqref{eq:ocp_formulation} with the given framework are discussed in \Cref{sec:results}.

%% file: tex/40_path_planning.tex
\section{Collision-Aware Path Planning}
\label{sec:path_planning}
Path-following MPC requires task-space references that are feasible with respect to obstacles and can be updated at a fast control rate.
%Path-following MPC requires feasible task-space references concerning obstacles and can be updated at a fast control rate. 
We employ a lightweight stochastic ensemble method for quadratic B-splines in the $4$-D task space $(x,y,z,\psi)$. 
The approach is inspired by spline-based motion planning methods \cite{NgN21,TiM18} and stochastic sampling strategies related to MPPI and Cross-Entropy methods \cite{2017Mppi, de2005tutorial}. In contrast to prior GPU-based implementations \cite{VuE25}, the proposed method is realized as a CPU-optimized implementation with OpenMP parallelization, targeting deployment in real-time control systems and achieving per-cycle runtimes of only a few milliseconds.

% \subsection{Anytime Planning Across Cycles}
The planner operates in an iterative manner: Per planner iteration, an ensemble of path candidates is sampled from a distribution, the best candidate is selected, and the distribution is updated. 
A single planner iteration does not need to converge; instead, quality improves across successive cycles by warm-starting from the previous distribution and best path. 
We use “anytime” in this specific sense: \emph{bounded computation per planner iteration, with improving solutions across iterations.}

\subsection{Problem Formulation}
Let $\vec{\xi}(s;\vec{c})\in \mathbb{R}^4$ denote a quadratic B-spline curve with clamped knots and control points $\vec{c}$ (for brevity, write \( \vec{\xi}(s) := \vec{\xi}(s;\vec c) \)). 
Start and goal are fixed, $\vec{\xi}(0)=\vec{\xi}_\text{s}$, $\vec{\xi}(1)=\vec{\xi}_\text{e}$, while interior via points $\{\vec{\xi}_{\text{v},i}\}_{i=1}^K$ adapt to avoid collisions and shorten the path. 
The control points are reconstructed by spline interpolation through the start, interior, and goal points. 
% We use $K{=}1$ interior via point (three via points in total), which suffices for the considered assembly tasks while keeping the optimization lightweight.

\subsection{Ensemble Sampling and Distribution Update}
In each planner iteration, an ensemble of candidate paths $\mathcal{P}=\{\vec{\xi}^{(m)}\}_{m=0}^{M}$ from $\vec{\xi}_\text{s}$ to $\vec{\xi}_\text{e}$ is generated by sampling interior via points from the current distribution,
\begin{equation}
  \vec{\xi}_{\text{v},i}^{(m)} \sim \mathcal{N}(\vec{\mu}_i,\ \vec{\Sigma}_i), \qquad i=1,\dots,K,
\end{equation}
with the following constraints: $(x,y,z)$ are truncated to workspace bounds, $z$ is projected above $z_{\min}$, and yaw $\psi$ is wrapped into the valid angular range. 
%If repeated Gaussian draws fall outside the bounds, a uniform sample is used. 
Together with the fixed start $\vec{\xi}_{\text{v},0}=\vec{\xi}_\text{s}$ and goal $\vec{\xi}_{\text{v},K+1}=\vec{\xi}_\text{e}$, these points define a candidate spline $\vec{\xi}^{(m)}$.
For the initial planner iteration (cold start) the mean value $\vec{\mu}_0$ is computed as linear interpolation from $\vec{\xi}_\text{s}$ to $\vec{\xi}_\text{e}$ and $\vec{\Sigma}_i = \sigma_\text{init}^2 \II$ with the parameter $\sigma_\text{init}$.

% \subsection{Evaluation and Cost Function}
Each candidate $\vec{\xi}^{(m)}$ is discretized at $N_c$ equidistant samples $\{s_j\}$. 
At each path pose $\vec{\xi}^{(m)}(s_j) = \transvec{\transpose{(\vec{p}^{(m)}_j)}, \psi^{(m)}_j}$ we compute: arc length increment $\norm{\vec{p}^{(m)}_j-\vec{p}^{(m)}_{j-1}}$, collision cost $d(\vec{\xi}^{(m)}_j)$, and a floor penalty
\[
P_\text{floor}(\vec{p}^{(m)}_{j}) = \begin{cases}
\alpha \left(z_\text{min}+\delta - z_j^{(m)}\right)^2, & z_j^{(m)} < z_\text{min}+\delta, \\ 0 & \text{otherwise}.
\end{cases}
\]
The composite cost is
% \begin{equation}
\begin{align}
J(\vec{\xi}^{(m)}) = \sum_{j=1}^{N_c-1} &\norm{\vec{p}^{(m)}_j-\vec{p}^{(m)}_{j-1}} \nonumber \\
&+ w_\mathrm{c} \big( d(\vec{\xi}^{(m)}_j) + P_\text{floor}(\vec{p}^{(m)}_{j}) \big) \Comma
\end{align}
% \end{equation}
with the factor $w_\mathrm{c} > 0$.
The collision cost $d(\vec{\xi}^{(m)}_j)$ is computed with Mujoco's GJK/EPA collision pipeline \cite{todorov2012mujoco}. 
%Candidates are ranked by $J$, and only collision-free ones are eligible as MPC references.

% \subsection{Distribution Update}
After evaluation, elites are selected by cost $J$ with normalized log-weights 
\begin{equation*}
     w_m \;=\; 
     \frac{\ln\bigl(M + 0.5\bigr) \;-\; \ln(m)}
          {\sum_{\nu=1}^{M}
            \Bigl[\ln\bigl(M + 0.5\bigr) \;-\; \ln(\nu)\Bigr]} \FullStop
   \end{equation*}
and for each via index $i$, the mean and variance are updated (warm start) as
\begin{align}
\vec{\mu}_i &\leftarrow \sum_{m\in\mathcal{E}} w_m \vec{\xi}_{\text{v},i}^{(m)}, \nonumber \\
\vec{\Sigma}_i &\leftarrow \sum_{m \in\mathcal{E}} w_m 
(\vec{\xi}_{\text{v},i}^{(m)} - \vec{\mu}_i) \transpose{(\vec{\xi}_{\text{v},i}^{(m)} - \vec{\mu}_i)} + \epsilon \II
\end{align}
where $\epsilon \II$ provides numerical stability.
%applied elementwise and clamped to $[\sigma_{\min},\sigma_{\max}]$. 
Exploration is adapted multiplicatively: if collision-free candidates exist, $\vec{\Sigma}_i$ contracts; otherwise, it expands. 
The reference path for MPC is the feasible candidate with minimal $J$.

\subsection{Benchmark Results}
\begin{figure}[t]
    \centering
    \begin{subfigure}[t]{0.48\linewidth}
        \centering
        \includegraphics[width=\linewidth]{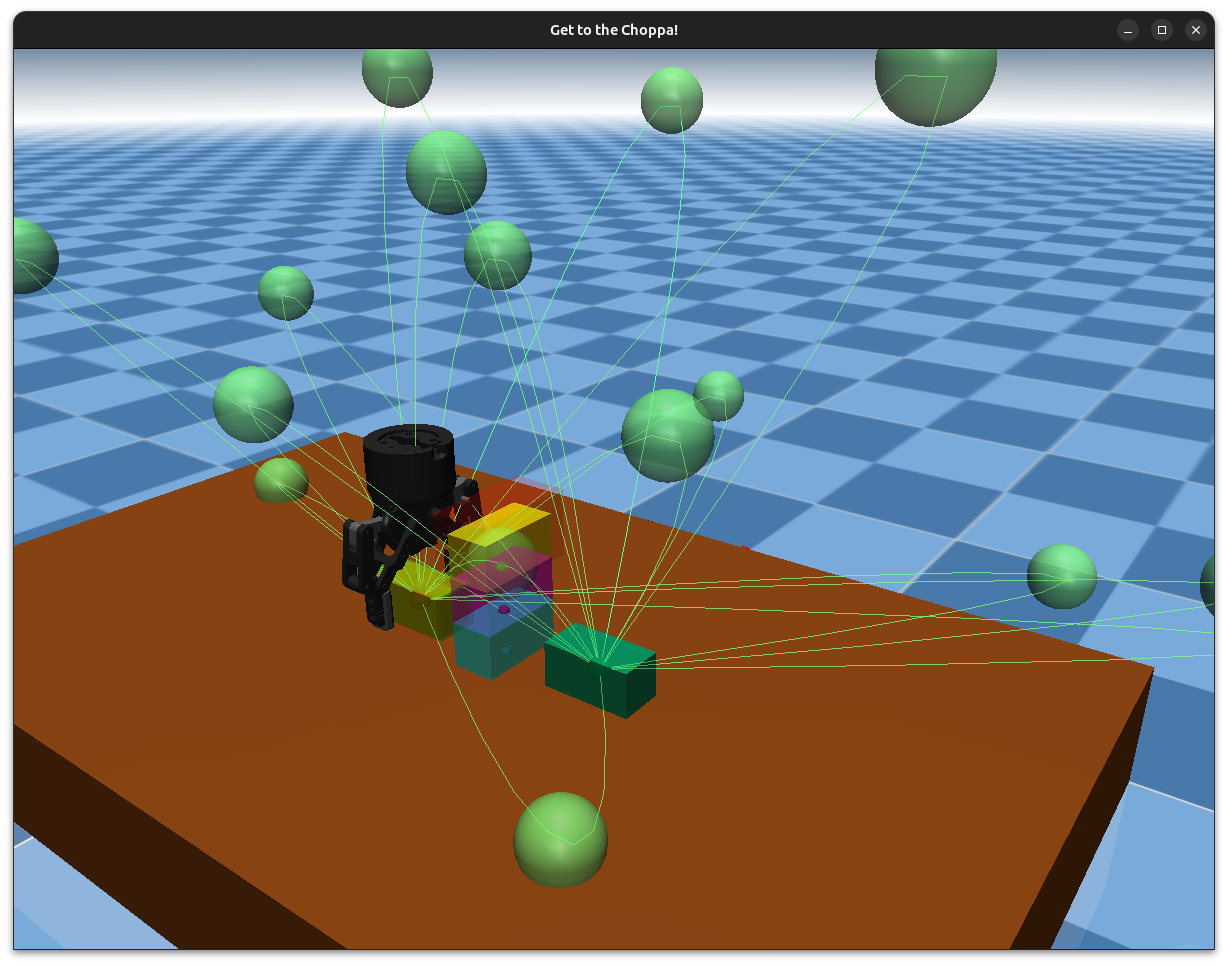}
        \caption{Initial ensemble sampled around the straight-line initialization.}
        \label{fig:ensemble_init}
    \end{subfigure}\hfill
    \begin{subfigure}[t]{0.48\linewidth}
        \centering
        \includegraphics[width=\linewidth]{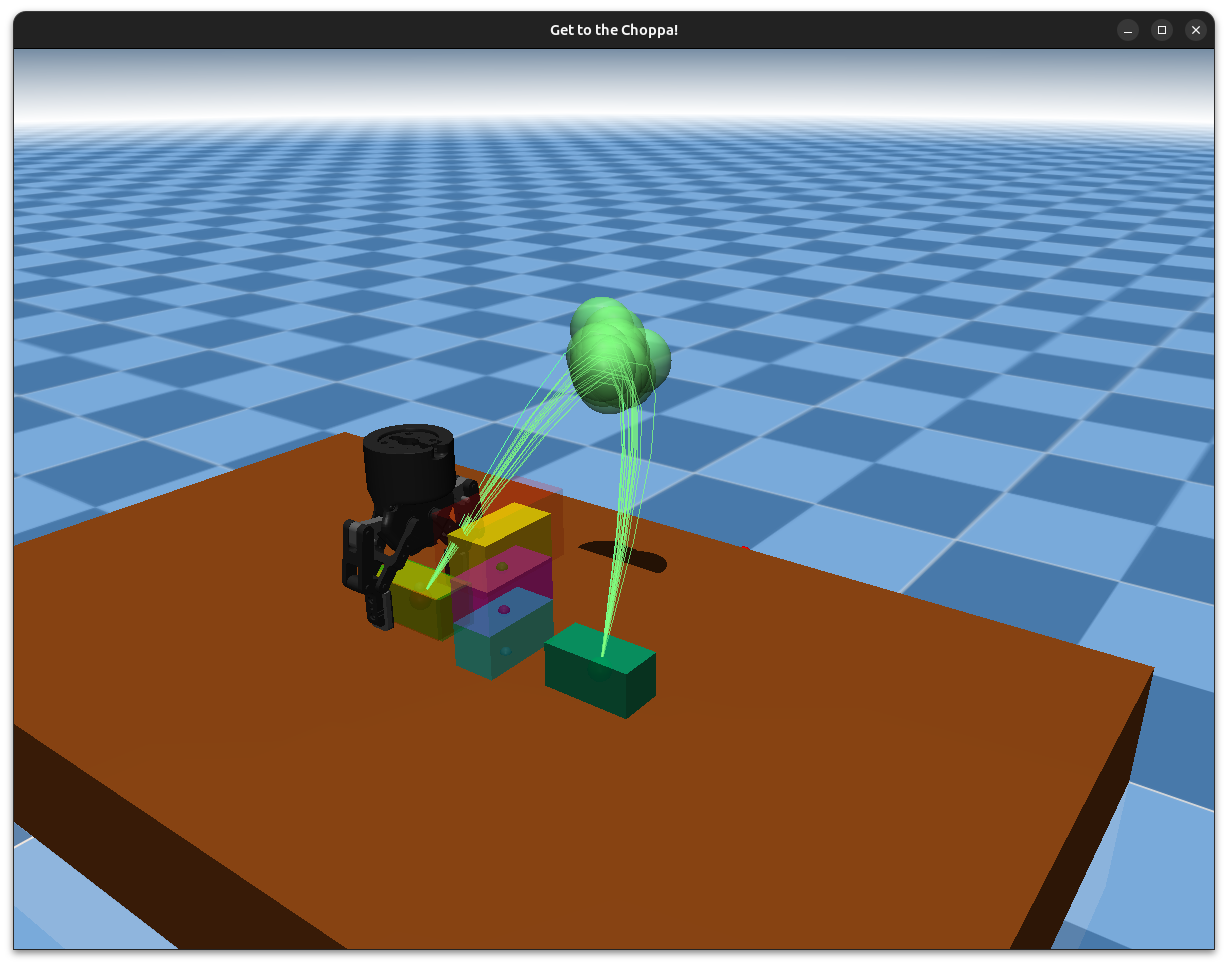}
        \caption{Ensemble after 10 refinement iterations.}
        \label{fig:ensemble_iter10}
    \end{subfigure}
    \caption{Evolution of the stochastic ensemble (15 candidates) with an obstacle wall of three stacked blocks. 
    Candidates start with wide variability (\subref{fig:ensemble_init}) and contract toward a feasible collision-free region (\subref{fig:ensemble_iter10}).}
    \label{fig:ensemble_evolution}
\end{figure}

The planner was benchmarked on the scenario depicted in \Cref{fig:ensemble_evolution} ($4$-DoF, one interior via point) over $N{=}100$ runs. 
We evaluate the planner with different target runtimes. Each run is stopped after the target time has elapsed; however, we allow the current iteration to finish before reporting. Thus, the reported times may slightly exceed the nominal target.
\Cref{tab:ces_bench} reports results for target runtimes $10,20,50$\, \si{\milli\second}. 
The first cycle is a cold start from the linear initialization; subsequent cycles continue with a warm-started distribution and the best path. 
Even under tight runtimes, the planner reliably produces feasible references, while larger runtimes yield shorter paths and higher iteration counts.

\begin{table}[ht]
\centering
\caption{Planner benchmark results ($N{=}100$ runs). 
Cold: first cycle from linear initialization; Warm: continued refinement with warm start. 
Runtime is per cycle. 
All runs use ensemble size $M{=}15$, one interior via point ($K{=}1$), initial standard deviation $\sigma_\text{init}{=}0.2$, 
and $N_c{=}40$ collision checks per path.}
\label{tab:ces_bench}
\begin{tabular}{llcccc}
\toprule
Target & Mode & Success & Time & Iter. & Length \\
\midrule
\SI{10}{\milli\second} & Cold & 95\% & \SI{12.6 \pm 2.9}{\milli\second} & 2.00 & \SI{0.780}{\meter} \\
                       & Warm & 94\% & \SI{12.5 \pm 3.4}{\milli\second} & 2.23 & \SI{0.753}{\meter} \\
\midrule
\SI{20}{\milli\second} & Cold & 98\% & \SI{23.1 \pm 2.4}{\milli\second} & 3.99 & \SI{0.685}{\meter} \\
                       & Warm &100\% & \SI{22.7 \pm 2.7}{\milli\second} & 5.41 & \SI{0.598}{\meter} \\
\midrule
\SI{50}{\milli\second} & Cold &100\% & \SI{53.6 \pm 3.9}{\milli\second} &13.45 & \SI{0.497}{\meter} \\
                       & Warm &100\% & \SI{53.3 \pm 3.7}{\milli\second} &12.66 & \SI{0.521}{\meter} \\
\bottomrule
\end{tabular}
\end{table}

%% file: tex/54_pose_filtering.tex
\section{Pose Estimation}
\label{sec:pose_estimation}
Using fiducial markers, our vision pipeline estimates the pose of blocks from RGB images $\mathcal{I}$. 
Although the generality of marker-based approaches is limited, they provide a simple and computationally efficient solution suitable for controlled laboratory validation under real-time constraints.

Given a calibrated camera image $\mathcal{I}$ with intrinsics $\vec{K}$ and distortion parameters, 
we detect ArUco markers $\mathcal{M}$ \cite{Aruco} and estimate their pose $\vec{H}_\mathcal{C}^\mathcal{M}$ (homogeneous transformation) using the IPPE PnP algorithm \cite{IPPE}.
With the fixed hand–eye calibration $\vec{H}_\mathcal{T}^\mathcal{C}$ and the robot forward kinematics $\vec{H}_{0}^\mathcal{T}(\vec{q})$, 
the marker pose in the robot base frame is obtained as
\begin{equation}
\vec{H}_{0}^\mathcal{M} = \vec{H}_{0}^\mathcal{T}(\vec{q}) \vec{H}_\mathcal{T}^\mathcal{C} \vec{H}_\mathcal{C}^\mathcal{M} \Rightarrow\;
\big(
\vec{p}_{0}^{\mathcal{M}}, \vec{r}_{0}^{\mathcal{M}} \big) \Comma
\end{equation}
with position $\vec{p}_{0}^{\mathcal{M}} \in \mathbb{R}^3$ and orientation $\vec{r}_{0}^{\mathcal{M}} \in \mathbb{H}$ (unit quaternion).

Since the block motion is unknown, we adopt a constant-velocity motion prior as a generic assumption. 
The position $\vec{p}_{0}^{\mathcal{M}}$ is filtered with a discrete Kalman filter on the state 
$\hat{\vec{x}}_{p,k}=\transvec{\vec{p}_{0,k}^{\mathcal{M}},\,\vec{v}_{0,k}^{\mathcal{M}}}$, 
while the orientation $\vec{r}_{0}^{\mathcal{M}}$ is smoothed using a multiplicative EKF (MEKF) on $\mathrm{SO}(3)$ with quaternion error representation \cite{MEKF,SO3}. 
Both filters run at the camera rate using the measured frame interval $\Delta t_k$.
The filtered 6D poses are fed into the world model (see Fig.~\ref{fig:system_overview}) 
and serve as goal references $\vec{\xi}_\text{e}$ for the path planner, which updates the path executed by the MPC.

%% file: tex/70_validation.tex
\section{Laboratory Setup}
\label{sec:laboratory_setup}
The experimental platform comprises a KUKA LBR iiwa 14 R820 equipped with a Robotiq 2F-85 parallel-jaw gripper and two passive joint DoFs. 
The unactuated coordinates $q_8$ and $q_9$ correspond to the relative roll and pitch between the robot end-effector and the gripper frame. 
These angles are inferred from a pair of IMUs (ADIS16460, Analog Devices) rigidly mounted on the end-effector and the gripper. Raw angular velocities are first debiased; orientation estimates are then obtained using a Madgwick filter \cite{MaH11a}. Let $\vec{R}_{\mathcal{W}}^\text{I1}$ and $\vec{R}_{\mathcal{W}}^\text{I2}$ denote the orientation (rotation matrices) of the end-effector-mounted and gripper-mounted IMUs in the inertial frame. The relative rotation is computed as
$\vec{R}_\text{I1}^\text{I2} = \transpose{ \left( \vec{R}_{\mathcal{W}}^\text{I1} \right) } \vec{R}_{\mathcal{W}}^\text{I2}$, 
from which the roll and pitch components are extracted to define $q_8$ and $q_9$. The IMU sampling rate is \SI{500}{\hertz}. An Intel RealSense D405 depth camera is rigidly co-located with the robot's end-effector and operated at \SI{15}{\hertz} for close-range perception and ArUco-based pose estimation.

All device drivers for the IMUs and the camera run on a Raspberry~Pi~4 (Ubuntu, ROS~2). The Madgwick orientation filter, ArUco pose filtering, task sequencer, path planner, and the model predictive controller (MPC) execute on a workstation (Ubuntu~22.04, Intel Core i7-12700K, \SI{32}{\giga\byte} RAM) running ROS~2 with Eclipse Cyclone DDS as middleware. Hard real-time joint control is deployed on a Beckhoff TwinCAT PC at $T_\text{c}=\SI{125}{\micro\second}$ (\SI{8}{\kilo\hertz}); ROS~2 and TwinCAT communicate via UDP, and TwinCAT interfaces with the robot over EtherCAT.

% \begin{table}[h]
%     \centering
%     \caption{MPC weights used in experimental validation.}
%     \label{tab:weighting_matrices}
%     \begin{tabular}{cc}
%         \toprule
%         \textbf{Weight} & \textbf{Value} \\
%         \midrule
%         $\vec{Q}_{\text{pos}}$ & $200 \II_3$  \\
%         $Q_{\text{rot}}$ & $1000$ \\
%         $\vec{Q}_{\vec{q}}$ & $\text{diag}([0.01, 0.01, 0.5, 0.01, 0.5, 0.01, 0.01, 0.01, 0.01])$ \\ 
%         $\vec{Q}_{\dot{\vec{q}}}$ & $\text{diag}([0.1, 0.1, 1.0, 0.1, 1.0, 0.01, 0.01, 1.0, 1.0])$ \\
%         $Q_s$ & $0.3 \eta$ \\
%         $Q_{\dot{s}}$ & $5$ \\
%         $\vec{R}_{\vec{u}}$ & $0.1 \II_7$ \\
%         $R_v$ & $5$ \\
%         \bottomrule
%     \end{tabular}
% \end{table}
% Table \Cref{tab:weighting_matrices} lists the weights in \eqref{eq:lagrange_cost} used during the experiments.
The weights in \eqref{eq:lagrange_cost} were tuned empirically in experiments, with high values on position and orientation tracking ($\vec{Q}_{\text{pos}}, Q_{\text{rot}}$) to ensure accuracy, and small posture and velocity regularization terms to stabilize redundancy, thereby balancing path-tracking precision, sway damping, and control effort in the experimental setup. 
The horizon length in \eqref{eq:ocp_formulation} is chosen as $N = 25$ steps, and the time discretization of the horizon is $T_\text{h} = \SI{30}{\milli\second}$, which is fast enough for the passive joint dynamics and long enough horizon for smooth path-following.
The MPC runs with $T_\text{s}=\SI{30}{\milli\second}$ and publishes a sequence of joint-accelerations $\vec{u}_{i|n} \equiv \ddot{\vec{q}}_{\text{a}}(t_{i|n})$ with $t_{i|n} = n T_\text{s} + i T_\text{h}$ and $i=0,\ldots,N$. In this work, real-time refers to the property that all MPC computations are completed within the fixed sampling interval $T_\text{s}$.

Since the MPC and torque control loops of the robot operate at different cycle times 
($T_s=\SI{30}{\milli\second}$ vs.\ $T_c=\SI{125}{\micro\second}$), 
signal consistency requires explicit up- and down-sampling. Linear interpolation of accelerations ensures that the inner torque loop receives continuous, cycle-synchronous references.
The robot is controlled by a computed-torque (CT) controller \cite{SiS09} with a sampling period $T_\text{c} = \SI{125}{\micro\second}$. To obtain the reference joint state and velocity at the control rate, the acceleration sequence $\vec{u}_{i|n} \equiv \ddot{\vec{q}}_{\text{a}}(t_{i|n})$ is first interpolated and then integrated with the semi-implicit Euler method.
The parameters for the computed-torque controller are obtained from \eqref{eq:eq_of_motion} with the mass of the passive joint and of the gripper included in the last link. 

A lightweight world model maintains the state of movable workpieces, including their estimated 6D poses and assembly status. 
This world model parametrizes a MuJoCo simulation updated online and queried for collisions during path planning (see Fig.~\ref{fig:system_overview}). 
The simulation provides the collision environment that the planner uses, ensuring that planning queries remain consistent with the real robot and workpieces. 
This way, perception, planning, and control are tightly coupled through a unified scene representation.

\section{Experimental Validation}
\label{sec:results}

\begin{figure}[htbp]
    \centering
    \begin{subfigure}[b]{0.35\columnwidth}
        \centering
        \includegraphics[width=\columnwidth]{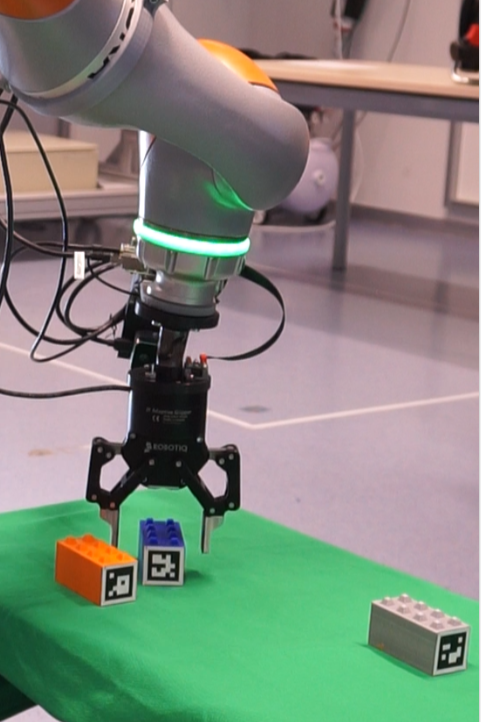}
        \subcaption{Pickup.}
        \label{fig:exp_pickup}
    \end{subfigure}
    \hspace{0.05\columnwidth}
    \begin{subfigure}[b]{0.35\columnwidth}
        \centering
        \includegraphics[width=\columnwidth]{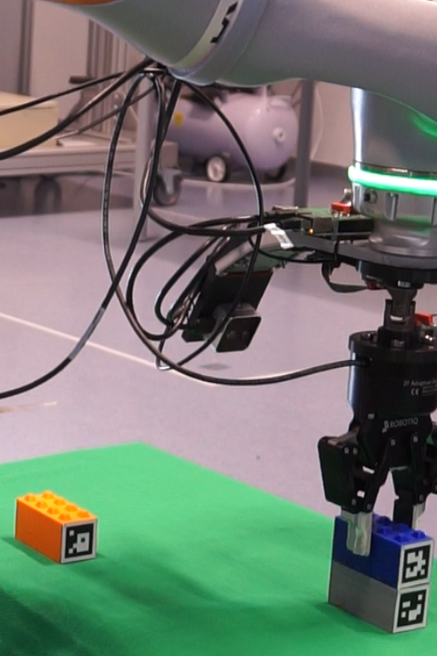}
        \subcaption{Placement.}
        \label{fig:exp_placement}
    \end{subfigure}
    \caption{Block pickup and placement. 
    In (\subref{fig:exp_pickup}), a randomly placed block is grasped using vision-based pose estimation feedback. 
    In (\subref{fig:exp_placement}), the block is placed on top of an existing block at a predefined goal, consistent with nominal assembly targets.}
    \label{fig:experiment_pickup_placement}
\end{figure}

The framework was validated in three experiments: The first and second experiments illustrate block pickup and placement tasks with online vision updates, and the third experiment shows the sway-damping performance of the presented method. 
% During execution, the vision system provides pose updates at \SI{1}{\second} intervals, which are fused into the world model and passed as updated goal references to the planner. 
For the first and second experiments, the planner iterations are executed at \SI{1}{\second} intervals. This interval was chosen to make replanning events clearly visible in the plots. The architecture supports faster updates limited by perception and computational constraints.
% to expose the effect of replanning and pose corrections clearly. 
%The lower limit of the time budget for the planner is determined by $T_\text{s}$, as faster updates would not have any effect.
% In practice, the underlying camera and filtering pipeline operate at \SI{15}{\hertz}, allowing faster updates if required.
In all experiments, we use $K=1$, \ie we have the start, goal, and one interior via point.
This minimal choice is sufficient for collision avoidance in the considered assembly scenarios while keeping the optimization lightweight. 
Determining the optimal number of via points for more complex tasks is an open question and left for future work. However, it is important to note that the proposed method is suitable for an arbitrary number of via points as it scales linearly with $K$, see \cite{ros2008simple}. 

\begin{figure}[htb] % or [h], [b], depending on placement
    \centering
    \includegraphics[]{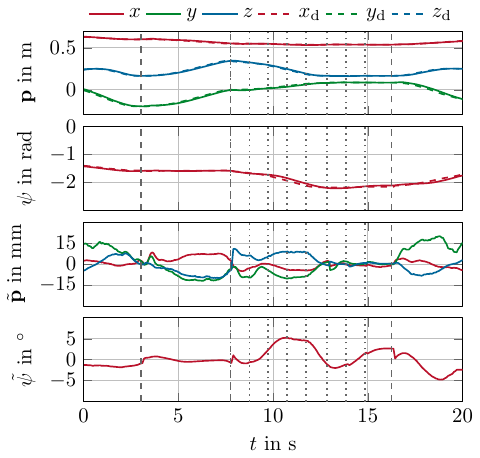}%
    \caption{Reference tracking and error during block pickup and placement (see \Cref{fig:experiment_pickup_placement}). 
    Vertical dashed lines indicate replanning events (new path initialization), while dotted lines mark vision updates (goal corrections from pose estimation). 
    From top to bottom: Cartesian position $\vec{p}$, yaw angle $\psi$, Cartesian position errors $\tilde{\vec{p}}$ and yaw angle error $\tilde{\psi}$. %components $x$ (red), $y$ (green), and $z$ (blue), with solid lines showing executed motion and dashed lines the reference path. 
    % Bottom: yaw angle $\psi$, executed (solid) vs.\ reference (dashed).
    }
    \label{fig:pose_plot}
\end{figure}

In the first experiment, depicted in \Cref{fig:experiment_pickup_placement}, blocks are picked up from random poses and stacked on top of each other.
Figure~\ref{fig:pose_plot} compares the reference trajectory (dashed) with the executed motion (solid), with replanning events marked by vertical dashed lines. 
The corresponding errors are shown in Fig.~\ref{fig:pose_plot}, remaining below \SI{3}{\milli\metre} and \SI{3}{\degree} during slow approach phases, while transient deviations up to \SI{15}{\milli\metre} and \SI{5}{\degree} occur during faster motions (e.g.\ around $t=\SI{11}{\second}$). 
These values remain within the tolerances required for block stacking in the laboratory setup.

\begin{figure}[htb] % or [h], [b], depending on placement
    \centering
    \includegraphics[]{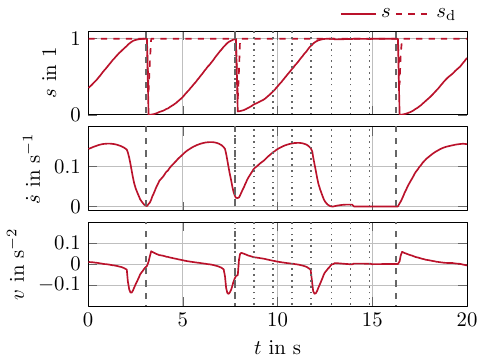}%
    \caption{Path progress during block pickup and placement. 
    Top: path parameter $s$. 
    %(solid: executed, dashed: reference; the reference ramps linearly from $0$ to $1$ for each new path to avoid discontinuities). 
    Middle: time derivative $\dot{s}$. 
    Bottom: input $v$.
    %, the additional control input of the augmented MPC formulation in \Cref{sec:control}.
    }
    \label{fig:path_progress}
\end{figure}

\begin{figure}[htb] % or [h], [b], depending on placement
    \centering
    \includegraphics[]{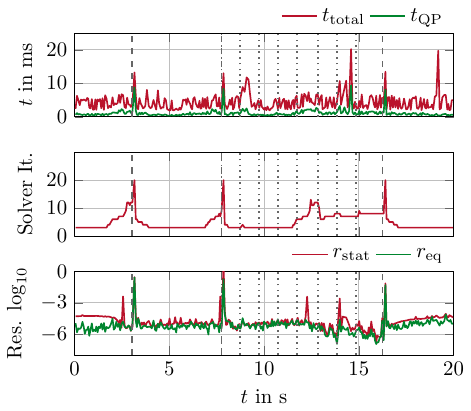}%
    \caption{MPC performance metrics during block pickup and placement. 
    Top: computation times, with $t_\text{total}$ the overall MPC iteration time and $t_\text{QP}$ the QP solver time. 
    Middle: number of QP solver iterations per MPC step. 
    Bottom: convergence residuals, where $r_\text{stat}$ measures first-order optimality (gradient norm) and $r_\text{eq}$ the satisfaction of system dynamics constraints.}
    \label{fig:acados_stats}
\end{figure}

The evolution of the path parameter $s$, its time derivative $\dot{s}$, and $v$ given in \Cref{fig:path_progress} illustrates how the path-following MPC advances along the spline under real-time updates. 
Solver performance is summarized in Fig.~\ref{fig:acados_stats}: QP solve times remain below \SI{20}{\milli\second}, iteration counts stay moderate, and residuals confirm stable convergence of the RTI scheme. 
Together, these results demonstrate that the integrated framework achieves real-time feasibility on CPU hardware while maintaining sufficient accuracy for autonomous assembly.

% \begin{figure}[htb] % or [h], [b], depending on placement
%     \centering
%     \includegraphics[]{graphics/experiments/pose_error_plot.pdf}%
%     \caption{Pose error during block pickup and placement. 
%     Top: Cartesian position errors 
%     %$\tilde{x}$ (red), $\tilde{y}$ (green), and $\tilde{z}$ (blue). 
%     Bottom: yaw angle error $\tilde{\psi}$. 
%     %All errors are computed as the difference between executed and reference trajectories shown in \Cref{fig:pose_plot}.
%     }
%     \label{fig:pose_error}
% \end{figure}

\begin{figure}[htbp]
    \centering
    % Erste Reihe
    \begin{subfigure}[b]{0.3\columnwidth}
        \centering
        \includegraphics[width=\columnwidth]{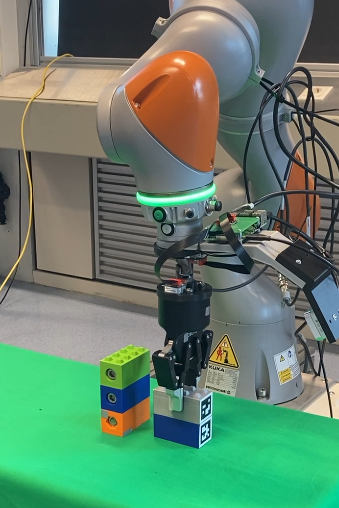}
        \caption{First pick up.}
        \label{fig:img1}
    \end{subfigure}
    \hfill
    \begin{subfigure}[b]{0.3\columnwidth}
        \centering
        \includegraphics[width=\columnwidth]{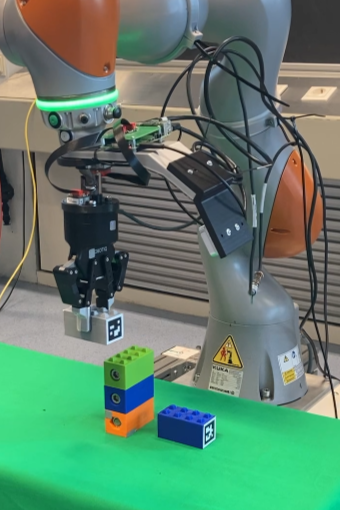}
        \caption{Traverse.}
        \label{fig:img2}
    \end{subfigure}
    \hfill
    \begin{subfigure}[b]{0.3\columnwidth}
        \centering
        \includegraphics[width=\columnwidth]{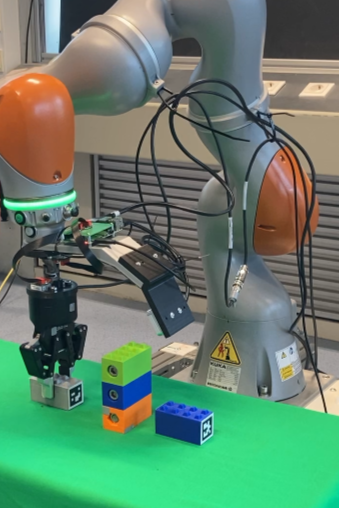}
        \caption{First placement.}
        \label{fig:img3}
    \end{subfigure}
    
    \vspace{0.5cm} % Abstand zwischen den Reihen
    
    % Zweite Reihe
    \begin{subfigure}[b]{0.3\columnwidth}
        \centering
        \includegraphics[width=\columnwidth]{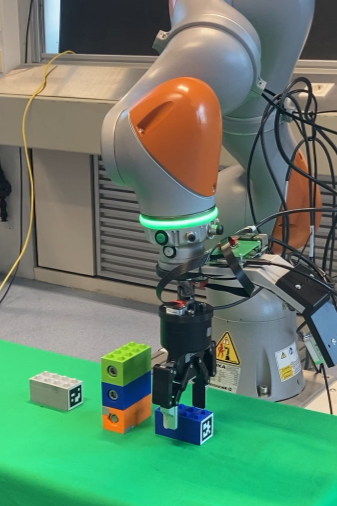}
        \caption{Second pick up.}
        \label{fig:img4}
    \end{subfigure}
    \hfill
    \begin{subfigure}[b]{0.3\columnwidth}
        \centering
        \includegraphics[width=\columnwidth]{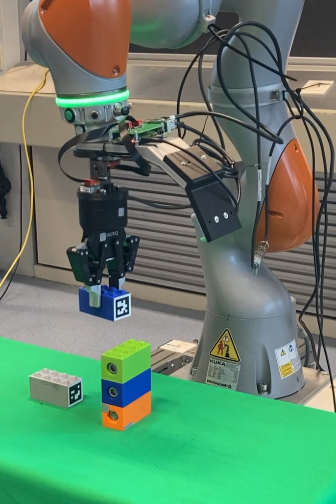}
        \caption{Traverse.}
        \label{fig:img5}
    \end{subfigure}
    \hfill
    \begin{subfigure}[b]{0.3\columnwidth}
        \centering
        \includegraphics[width=\columnwidth]{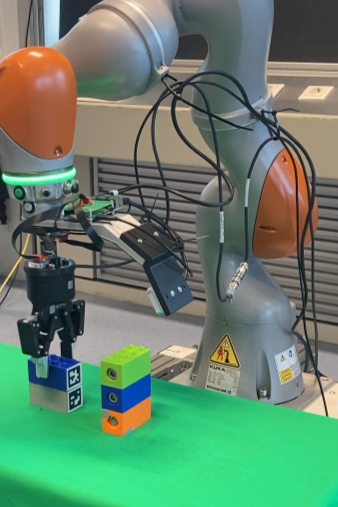}
        \caption{Sec. placement.}
        \label{fig:img6}
    \end{subfigure}
    
    \caption{Assembly of blocks with an obstacle wall (see \Cref{fig:ensemble_evolution}). 
    The wall consists of three stacked blocks separating pickup and placement locations. 
    Two blocks are transferred across the obstacle in sequence, each consisting of pickup, traverse, and placement.}
    \label{fig:obstacle_avoidance}
\end{figure}

In the second experiment, blocks are picked up on one side of an obstacle wall and stacked on the opposite side. \Cref{fig:obstacle_avoidance} illustrates the experiment sequence. The accompanying video provides a clearer view of this experiment, particularly the gripper re-adjustments during pickup based on vision feedback.

\begin{figure}[htb] % or [h], [b], depending on placement
    \centering
    \includegraphics[]{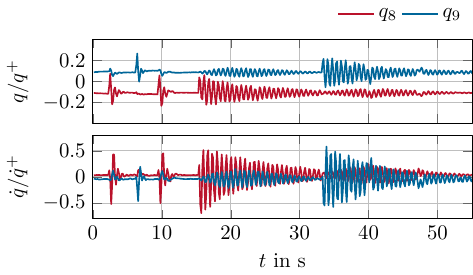}%
    \caption{Damping performance of the MPC controller under external excitation. 
    The controller is active from $t=\SI{0}{\second}$ to $t=\SI{15}{\second}$, after which the natural, lightly damped sway motion of the passive joints becomes visible. 
    Quantitative parameters of the fitted oscillations are reported in \Cref{tab:sway_damping}.
    }
    \label{fig:mpc_damping}
\end{figure}

% \subsection{Sway Damping}
The third experiment depicted in \Cref{fig:mpc_damping} illustrates the damping of passive joint oscillations achieved by the MPC controller. The gripper was manually perturbed to excite the sway angles $q_8$ and $q_9$ to evaluate performance. The controller was active from \SI{0}{\second} to \SI{15}{\second} and then deactivated, exposing the natural dynamics of the under-actuated joints. A damped-sine fit was applied to both cases, and the key parameters are summarized in \Cref{tab:sway_damping}.
\begin{table}[htb]
    \centering
    \caption{Damping characteristics of $q_9$ with MPC active vs.\ inactive. 
    With MPC active, the decay rate and damping ratio increase by more than an order of magnitude, 
    reducing the settling time from \SI{17.8}{\second} to below \SI{1}{\second}.}
    \label{tab:sway_damping}
    \begin{tabular}{lccc}
        \toprule
        & Decay rate $\sigma$ & Damping ratio $\zeta$ & $T_{90\to10}$ \\
        \midrule
        MPC on  & \SI{2.50}{\per\second} & 0.23  & \SI{0.88}{\second} \\
        MPC off & \SI{0.12}{\per\second} & 0.013 & \SI{17.8}{\second} \\
        \bottomrule
    \end{tabular}
\end{table}
The results highlight the exceptional damping achieved by the MPC controller: the effective damping ratio increases from $\zeta=0.013$ (barely damped) to $\zeta=0.23$, reducing the settling time by a factor of more than 20 (from \SI{17.8}{\second} to less than \SI{0.9}{\second}). Notably, the oscillation frequency remains nearly unchanged (\SI{1.67}{\hertz} vs. \SI{1.54}{\hertz}), showing that the improvement is due to active damping rather than altered dynamics. This confirms that the MPC effectively suppresses passive joint sway, stabilizing the gripper motion in real time.
All experimental results are summarized in an accompanying video available at \expvideo.

%% file: tex/90_conclusion.tex
\section{Conclusion}
\label{sec:conclusion}
This paper presents a unified perception–planning–control framework for under-actuated manipulators with passive joints, realized on a laboratory-scale testbed designed to emulate articulated boom cranes. Integrating real-time vision feedback, collision-aware path generation, and nonlinear MPC enables closed-loop execution of assembly tasks under dynamic and cluttered conditions.  

Experiments on the testbed demonstrated autonomous block pickup, placement, and obstacle-avoidance assembly with online vision updates. The results confirmed both the real-time feasibility of the architecture and its ability to actively damp passive-joint sway, reducing settling times by more than an order of magnitude.  

The laboratory-scale setup proved to be an effective intermediate environment: simple enough for rapid prototyping and controlled experimentation, yet rich enough to capture the key challenges of under-actuated crane dynamics. This bridge between tabletop robotics and full-scale machinery allows systematic development of autonomy concepts under realistic but manageable conditions.  

Future work will extend the vision system beyond fiducial markers, explore richer assembly tasks, and investigate the transition to large-scale machinery.